# Application of Machine Learning in understanding plant virus pathogenesis: Trends and perspectives on emergence, diagnosis, host-virus interplay and management

Dibyendu Ghosh[1†], Srija Chakraborty [2†], Hariprasad Kodamana [2,3], Supriya Chakraborty[1*]

[1]Molecular Virology Laboratory, School of Life Sciences, Jawaharlal Nehru University, New Delhi-110067, India

[2]Department of Chemical Engineering, Indian Institute of Technology Delhi, New Delhi – 110016, India

[3]School of Artificial Intelligence, Indian Institute of Technology Delhi, New Delhi – 110016, India

## ORCID ID:-

Dibyendu Ghosh - 0000-0001-9707-0635

Srija Chakraborty - 0000-0001-7357-3713

Hariprasad Kodamana - 0000-0003-3166-2712

Supriya Chakraborty – 0000-0001-9301-2649

[†] Contributed equally

[*]Address for correspondence

Prof. Supriya Chakraborty

Molecular Virology Laboratory,

School of Life Sciences, Jawaharlal Nehru University,

New Delhi-110067, India

e-mail:-schakraborty@mail.jnu.ac.in, supriyachakrasls@yahoo.com

Telephone: 91-1-2670 4153

Fax numbers: 91-11-2674 2558

Dibyendu Ghosh : dibyendughosh48@gmail.com

Srija Chakraborty: srijaciitd@yahoo.com

Hariprasad Kodamana: kodamana@iitd.ac.in

**Total word count**  :   3779

**No. of figures**       :   3

**No. of tables**        :   1

**Abstract**

**Background:** Inclusion of high throughput technologies in the field of biology has generated massive amounts of biological data in the recent years. Now, transforming these huge volumes of data into knowledge is the primary challenge in computational biology. The traditional methods of data analysis have failed to carry out the task. Hence, researchers are turning to machine learning based approaches for the analysis of high-dimensional big data. In machine learning, once a model is trained with a training dataset, it can be applied on a testing dataset which is independent. In current times, deep learning algorithms further promote the application of machine learning in several field of biology including plant virology.

**Main body:** Plant viruses are emerging as one of the principal global threats to food security due to their devastating impact on crops and vegetables. The emergence of new viral strains and species help viruses to evade the concurrent preventive methods. According to a report conducted in 2014, plant viruses are anticipated to cause a global yield loss of more than thirty billion USD per year. In order to design effective, durable and broad-spectrum management protocols, it is very important to understand the mechanistic details of viral pathogenesis. Application of machine learning enables precise diagnosis of plant viral diseases at an early stage itself. Furthermore, the development of several machine learning guided bioinformatics platforms has primed plant virologists to better understand the host-virus interplay. In addition, machine learning has tremendous potential in deciphering the pattern of plant virus evolution and emergence and developing viable control options.

**Conclusions:** Considering a significant progress in the application of machine learning in understanding plant virology, this review highlights an introductory note on machine learning and comprehensively discusses the trends and prospects of machine learning in diagnosis of viral diseases, understanding host-virus interplay and emergence of plant viruses.

# Background

## Machine learning: an introduction for biologists

Over the years, extensive research has been carried out in various fields of biology to understand the science behind a plethora of complex biological phenomena. The study of problems such as traits in plants and plant viral diseases lead to generation of massive data sets. The progress in technology has rendered data generation a simple task. Cost-effective technologies such as Next Generation Sequencing (NGS) have made it easier to gather data regarding gene expression, chromosome conformation, genetic variation, traits and diseases of animals and plants leading to generation of such massive data sets having multiple characteristics [1]. However, the resultant data explosion, especially in the field of omics, has made the handling of large datasets a major concern. The traditional statistical data analysis methodologies are not effective or efficient anymore in this context [2].

Furthermore, biological phenomena comprise various aspects, which lead to the generation of more than one data type. This calls for an integrated analysis of the different types of data. But the noisiness of heterogeneous biological data makes this a difficult task [3]. Data dimensionality is another major impediment, for instance, omics data is generally highly resolved, hence highly dimensional. Moreover, sample size in biological studies is limited in most cases. This may lead to issues including overfitting, multi-collinearity and data sparsity [4].

In order to overcome all these barriers, attempts are being made to incorporate machine learning (ML) and deep learning (DL) tools in the analysis of the datasets. ML tools identify patterns in the data using different statistical methods. ML paradigm can be used to derive models for classification, pattern recognition, and making predictions based on existing data. DL algorithms extract high level features from huge datasets (like a collection of genomic sequences, or images), recognize the hidden patterns and then use them to create trained prediction models [5]. These trained models can be further applied to different data from other sources for applications such as prediction and classification. These techniques have the ability to tackle tough problems by seeing structure in seemingly random data, even when the amount of data is too complex and large for human comprehension [6]. Hence, ML especially DL has the ability to perform analysis of enormous datasets in an extremely efficient, cost-effective, accurate and high-throughput manner [7].

In the context of ML, there are two primary methods for training the models: supervised and unsupervised learning. Both of these have potential for use in biology. Under supervised learning, the given collection features, or attributes of a system under investigation are labeled [8]. Two recurring problems in the supervised learning framework are regression and classification. The classification process assigns objects into the classes on the basis on the properties of features. In biology, one example of such training (involving mapping of object-to-class) is mapping of gene expression profiles to their respective diseases. The algorithm returns a "doubt value", a parameter that shows how unclear the algorithm is about assigning the object to one of the many possible cases, or an "outlier", which shows how unlikely the object is to the other previous objects making classification tough. Some of the widely used supervised modes are linear/nonlinear regression, support vector machines (SVM), Gaussian processes, and neural nets [9].

In unsupervised learning, the objects involved in the study are not under any predefined labels [10]. The entire goal of these models is to recognize similarities in the various objects by exploring the data. These similarities define clusters in the data (groups of data objects). So, the basic concept of unsupervised learning is to discover natural patterns and form groupings of the data. Thus overall, in supervised learning, the data is pre-labeled and the algorithm learns how to use the labels to associate the objects to the classes. On the other hand, in unsupervised learning, the data is unlabeled, and the algorithm also learns to create labels by clustering the objects. Principal component analysis includes k-means clustering, Gaussian mixture models, Density-based spatial clustering of applications with noise (DBSCAN), and hierarchical clustering [9].

In certain exceptional cases, a method known as semi-supervised learning has proven to be quite useful. One example of such a situation could be the classification of protein sequences. Here, only a few samples of protein sequences are labeled (belonging to a known class) while quite many sequences belong to unknown classes. The semi-supervised algorithm transfers the class labels from the labeled to the unlabeled objects present in the feature space nearby [11].

The basic steps for creating a machine learning model for the study of biological data are shown in Figure 1. After gathering the data (labeled or unlabeled), it is split into two sets for training and then testing. The data samples need to undergo preprocessing and augmentation before the splitting in case they are corrupted with noise and outliers. Next the model is trained using the training dataset. The model can either be created from scratch, or a pre-trained model can be adjusted according to the dataset collected. Once the trained model is ready, the testing data is passed through it to check the accuracy with which the objects are classified into different labels [12].

When working with neural networks, we essentially attempt to create the inferences analogous to the human brain by building an Artificial Neural Network (ANN) [13]. An ANN resembles a biological neural network. The artificial neurons used here are basically mathematical models that carry out three main functions: activation, addition and multiplication. The goal is to build layers of neurons, each of which produces a suitable response to any input provided to it. The neurons of each layer multiply their inputs with the corresponding weights. Then it is passed through the activation function and finally transferred to the next layer of neurons. Once the input layer is fired up, the decision moves along to the layers of the neurons (hidden layers), firing up the respective neurons until the final output layer is reached [14]. A schematic representation of a neural network is shown in Figure 2.

In neural networks, the information flow direction is determined by the intermodal connections. On this basis, there are two classifications of neural networks: for unidirectional flow, we have cascade forward and feed-forward; for bidirectional flow we have feed-backward or recurrent [15]. In feed-forward networks, the flow of information between the layers takes place in one direction. Cascade forward is similar except the input to the next layer is weighted. In recurrent networks, flow of information takes place in both directions. All the nodes are interconnected among each other, including self-connection. These networks are unfortunately extremely complex, bulky, difficult to operate and take up a large amount of computational space. There are many other neural networks whose development is in progress, some including self-organizing networks, convolutional neural networks (CNN), variational auto encoders (VAE), generative adversarial networks (GAN) [16, 17].

Various parameters come in use while evaluating the classification performance of the model developed. Some of the important parameters include accuracy, sensitivity/recall rate, specificity rate, precision/positive predictive value, negative predictive value, $F1_{scor}$ [18]. These judge the performances of models by calculating various ratios involving true positives, false positives, true negatives and false negatives. All of these can be combined into a single confusion matrix, which is then studied to judge the model's performance [19]. In addition to this, the phenomena of overfitting and underfitting are widely faced while employing ML models and, therefore, are being investigated by researchers [20]. Overfitting is a situation when the fitting of the model is with respect to the noise in the data and not the signal. The validation data error increases while the training data error decreases [21]. On the other hand, underfitting is the reverse scenario. In this case, the model is not capable of recognizing data variability [22]. Many methods are being developed in an attempt to develop the perfect model and to prevent any such imperfect fittings. Penalty methods, training by early stopping, batch normalization, and dropouts are to name a few [23].

**Main text**

**Application of Machine Learning in understanding plant viruses and viral infections**

**I. Diagnosis and detection of plant viral diseases:**

Plant viruses pose the major economical constraints in cultivated crop plants across the world. Early detection of plant viral infection is crucial for successful disease management. The empirical evaluation through visible survey is traditionally followed by farmers to identify the symptoms of virus infected diseased plants. The visual assessment bias dictates the inefficiency and inaccuracy of this method. On the other hand, laboratory-based detection techniques are primarily reliant on polymerase chain reaction (PCR) and serological driven methods, such as enzyme linked immune sorbent assay (ELISA) . Despite their improved accuracy, the requirement for professional experts and their time-consuming and non-invasive nature pinpoints the shortcomings of these diagnosis assays [24, 25]. A pathogen attack significantly alters the biochemical and biophysical state of the plant leading to an alteration of tissue structure, water level and transpiration rate, ultrastructure of chloroplast and pigment content [26, 27]. At the very beginning of the 21st century, a few studies used remote sensors to capture and detect altered leaf reflectance and thermography profiling of diseased plants, which empowered the scientific community with an edge in phenotyping of stressed plants [28]. However, this technique was unable to determine if the stress was biotic or abiotic, and if biotic, what kind of pathogen was involved. Hyper-spectral imaging (HSI) and ML assisted data analysis are now revolutionizing the concept of stress phenotyping of diseased plants by enabling the diagnosis of specific plant diseases and even the severity of the disease. In the case of HSI, a light spectrum with a larger range of wavelengths is being used to capture plant images, which enables us to go beyond the limited range of human vision (400-700nm) in monitoring minor alterations in the growth and development of plants [29]. For ML assisted detection of plant viral diseases, a ML model has to be trained first with a training dataset (images of diseased plants captured through unmanned aerial vehicle, grounded robots or even smartphones) [30-37]. There are increasing numbers of free online databases which provide images of specific plant diseases as training datasets. 'Plantvillage' is one such initiative [38]. Once, a certain ML model has been trained accurately and precisely, a testing dataset (eg: hyperspectral images of specific plants under diagnosis) can be assessed [32, 34]. HSI generates high dimensional data with redundant information and hence, an efficient pre-processing of the data is crucial for the precise functioning of the model. An effective specific range of wavelength can be determined to reduce the dimensionality of HSI data [29]. The next step is feature extraction which minimizes the number of the features present in the raw dataset. The feature extraction method is vital for assuring a simple classifier with a limited variety of features, since multifeatured classification always hinders the smooth performance of the concerned model [39, 40]. ML researchers have devised a variety of feature extraction techniques based on the nature of the data and the model. However, the process is time-consuming and the success of the operation greatly relies on the expertise of the professional. Here comes the benefits of using DL techniques for feature extraction as DL empowers automatic extraction of features rather than handcrafted method used in traditional ML algorithms, for instance the application of convolutional neural networks. DL has substantially improved the reliability of plant stress phenotyping by enabling the accommodation of a large sample size for training and testing [39]. A major constraint of this method is the vast variation of environmental conditions between the field and the lab. While a consistent temperature, humidity, and light intensity are maintained in the lab, all of these variables are constantly changing in the field, influencing the captured images [41]. Hence, it is recommended to use field images to train a model since it has been demonstrated that a classifier trained on field images can also classify lab-based images with precision [42]. The lack of availability of a huge collection of field-based images of specific plant disease is another key challenge for an accurate and reliable diagnosis.

Transfer learning is a recent advancement in the field of ML which enables the data scientists to adopt a previously well-trained model for solving a similar kind of problems [39]. For example-a model trained for chilli-leaf curl disease detection may be used for detecting leaf curl symptoms caused by viruses in tomato (Figure 3). There are several approaches to adopt a pre-trained model; one can select and finetune the architecture and/or parameters of a model depending upon the types of datasets. **Table 1** summarizes the development of ML assisted diagnosis of plant viral diseases over last few years.

**II. Understanding the diversity and emergence of plant viruses:**

The recent trend of studying plant virome through metagenomics has unveiled the diversity of plant viruses. Huge numbers of phylogenetically related and unrelated virus species have been found in diseased samples [43, 44]. Havoc explosion of virome data generated through NGS necessitates the urgent structural orientation and analysis of sequence data in order to understand the actual portrait of the viral diversity. Although a significant progress has been followed up in the case of animal viruses, limited efforts have yet been recorded in the field of plant virology [45, 46]. V-pipe has provided a bioinformatics pipeline for analyzing genomic diversity of human immunodeficiency virus (HIV) from sequencing data[47].

As RNA viruses use error-prone polymerases during their replication, the chances of mutations in their genome sequences remain quite high. Mutation in the viral genome finally leads to the emergence of new virulent viral strains [48]. A neural network-based model can predict probable point mutations in the RNA sequence. It has been successfully explored in the case of newcastle virus [49] and its optimized form may be very useful for prediction of mutations in plant viral genome (Figure 3). Besides RNA viruses, DNA viruses also possess significant genetic variations and events like recombination and genome reassortment play crucial role in mediating the emergence of new viral forms [50, 51]. The identification of novel virus and satellite molecules through metagenomics approach emphasizes the importance of precise taxonomic classification followed by demarcation of these new species. An excellent effort by Silva and collaborators have developed Fangorn Forest, a ML based method, for classification of geminiviruses. Among the three tested algorithms, random forest (RF) has been proven to be best in classification of genes and genera of this largest plant virus family [52].

**III. Understanding host-virus interplay:**

Being obligate parasites, viruses rely on cellular machineries of plants for every aspect of pathogenesis including replication, gene expression and movement [53]. Plants elicit a robust antiviral immune response to restrict viral invasion [54]. Viruses encode effector proteins which disarm plant defense signaling. This tug of war continues which fuels the co-evolution of both virus and host [55]. Hence, understanding the interplay between plant and viruses is crucial for in-depth dissection of viral pathogenesis.

Although plants have evolved a variety of tools and tactics to prevent virus multiplication, the resistance (R) protein-mediated immune response and gene silencing are the most well-known features of their antiviral defense [54]. The majority of canonical R-proteins contain nucleotide binding site leucine-rich repeats (NBS-LRR), which mediate direct or indirect recognition of virus-encoded effector proteins, resulting in the activation of effector triggered immunity (ETI). Very few *R*-genes imparting immune response against viruses have been identified and characterised till date, which limits our knowledge regarding the detailed mechanism of dominant resistance in plant virus interaction [56]. Support vector machine-assisted development of a high throughput bioinformatics tool, NBSPred, precisely identifies NBS-LRR containing R proteins from genome, transcriptome and proteome data [57]. Receptor-like kinases (RLK) are crucial players in the immune perception of phytopathogens, many of them acting as pattern recognition receptors (PRRs) which lead to induction of pattern triggered immunity (PTI) [58]. However, several plant viruses target RLKs to promote viral pathogenesis [59]. Brustolini et al. have recently developed a machine learning assisted technique for detection of RLKs from proteome data. Identification and annotation of novel RLKs may advance our current understanding of plant-virus interactions [60]. Furthermore, to identify host factors differentially regulated in host- virus interplay, several groups have performed transcriptome, proteome analysis in both resistant and susceptible plant varieties and these studies have revealed that a significant proportion of differentially expressed transcripts are of unknown nature suggesting the existence of novel gene regulatory networks (GRNs) modulating the host-virus interaction [61-64]. ML helps biologists to predict GRNs from high-throughput transcriptome data [65] which may lead to identification of several regulatory nodes of plant immune signalling (Figure 3).

On the other hand, viruses encode few but multitasking viral effector proteins which facilitate the viral pathogenesis. Examining the sub cellular localization of these effector proteins is important to understand their mechanism of action. Furthermore, viruses also redirect the subcellular localization of several host proteins to disrupt their assigned functions [66]. ML assisted development of online tools such as LOCALIZER, MU-LOC enable precise as well as accurate analysis of subcellular localization of effector proteins and host factors by simply using amino acid sequences of proteins as input (Figure 3) [67, 68]. Application of ML in the successful prediction of fungal effector proteins has added an extra edge in phytopathology research [69]. In the case of viruses, some viral effector proteins have been evolved to block antiviral gene silencing, known as viral suppressors of RNA silencing (VSRs). VSRs expands the negative impact of viral diseases by promoting synergistic associations among different plant viruses [70]. Jagga et al have developed a bioinformatics platform, pVsupPred, for the prediction of VSRs encoded by plant-infecting viruses. They have used four classifier models including LibSVM, J48, Naïve Bayes, RF and among all of them, RF algorithm has emerged as the best with an overall accuracy of 86.11% [71]. Later on, in another study, sequential minimal optimization (SMO) algorithm had been optimised to achieve an overall accuracy of 95.3% for the successful identification of plant virus encoded VSRs (Figure 3) [72].

Another significant facet of plant-virus interaction is the virus induced alteration of microRNA (miRNA) homeostasis which impacts the transcriptome profile of the infected cells. Hence, it is important to identify the accurate targets of specific miRNAs regulating plant immunity and viral pathogenesis [73]. Advent of ML in advancing the scope of bioinformatics has significantly eased this difficult job. Supervised ML approaches including graphical models, kernel machines and evolutionary algorithms are being widely used to identify the specific miRNA targets in eukaryotes [74]. Further, a new category of DL models known as graph neural nets (GNN) is emerging as a promising tool in bioinformatics. The biological networks, based on small RNAs–disease associations can be constructed as graphs with nodes and edges. GNN can operate on the graphical data and has more representative features, which can be efficiently used for inferences [75].

Finally, the best possible way to understand the functional aspect of a protein is to visualize its accurate structure. A very small proportion of plant proteins involved in immune signalling have been structurally characterised yet. In addition, structures of plant viral proteins are also largely unresolved. Labour intensive methods of protein crystallization is the major bottleneck here. However, Jumper and collaborators have revolutionised the idea of protein structure prediction by launching Alphafold2, a neural network assisted structural bioinformatics platform, which can successfully solve a protein structure with almost equivalent experimental accuracy even if there is no similar protein structure available [76]. ML-guided docking studies efficiently screen chemical inhibitors of severe acute respiratory syndrome coronavirus 2 (SARS-CoV-2) encoded spike (S) protein [77]. Similarly, structure prediction of plant viral proteins and the prediction of their chemical inhibitors followed by successful delivery will be a novel and effective virus management strategy (Figure 3).

**Conclusions and future perspectives**

Although the last decade has witnessed a sharp increase of application of ML in solving complex biological problems [78], its usage in the field of plant virology is still at a very naïve state. Several reports highlighted the role of ML in the precise diagnosis of plant viral diseases [30-36]. Plant virologists can foresee the tremendous scope of ML in addressing virus evolution, emergence, plant-virus interplay and above all management strategies. Moreover, several specific issues need to be explored.

Firstly, the coordinated application of ML and HSI pave a new path in detection of viral diseases. Now, designing improved DL-based algorithms for easily accessible mobile app mediated detection of plant viral diseases is the need of the hour. Secondly, a vast number of OMICS data (including transcriptome, proteome, metabolome) of virus infected plants are available. Application of ML may enable the integration of these OMICS data which will definitely uplift our knowledge of host response in viral infection. Thirdly, ML algorithms would be very helpful in deciphering the patterns and parameters of plant viral evolution. ML may enable the accurate prediction of recombination, rate of nucleotide substitutions, mutations and phylogenetic relatedness among viruses. Fourthly, a significant improvement can be achieved in metagenomics data analysis through ML approaches. Along with viral reads, metagenomics data also contains a substantial proportion of host contig contaminations which often hinders the identification of small viral reads. VirFinder is one such *k*-mer based platform which enables the identification of prokaryotic virus sequences from mixed metagenomic data [79]. A similar approach can be employed for plant virome study. Fifthly, ML is now widely utilized in genomic selection for rapid and better prediction of superior genotypes for breeding purposes [80, 81]. A certain progress of ML assisted genomic prediction will definitely help breeders in developing elite virus tolerant/ resistant varieties. Finally, a collaborative effort from both plant virologists and big data analysts is of prime importance for the fruitful application of ML in the understanding of plant virus pathogenesis followed by the development of antiviral strategies.

**Abbreviations**

NGS : Next Generation Sequencing; ML: machine learning; DL: deep learning; SVM: support vector machines; DBSCAN: Density-based spatial clustering of applications with noise ; ANN: Artificial Neural Network; CNN : convolutional neural networks; VAE : variational auto encoders; GAN : Generative adversarial networks; PCR: Polymerase chain reaction; ELISA : enzyme linked immune sorbent assay; HSI: Hyper-spectral imaging; HIV: human immunodeficiency virus; RNA: Ribonucleic acid; DNA: Deoxyribonucleic acid; R protein - Resistance protein; NBS-LRR: nucleotide binding site leucine-rich repeats; ETI : effector triggered immunity; PTI: Pattern triggered immunity; RLK: Receptor-like kinases; PRRs : Pattern recognition receptors; GRNs: Gene regulatory networks; VSRs: Viral suppressors of RNA silencing; RF: Random forest; SMO: Sequential minimal optimization; miRNA: microRNA; GNN: Graph neural nets; SARS-CoV-2: Severe acute respiratory syndrome coronavirus 2; S protein : Spike protein; CNN : Convolutional neural networks; OR-AC-GAN : Outlier removal auxiliary classifier generative adversarial nets; SVM : Support vector machine; SPA : Successive projections algorithm; ELM : extreme learning machine; VGG 16 : Visual geometry group 16; BRT: Boosted regression tree; TSWV : Tomato spotted wilt virus; GBNV : Groundnut bud necrosis virus; TMV : Tobacco mosaic virus

## Acknowledgements
Not applicable.

## Authors' contributions
DG, SRC, HK and SC conceived the work; DG and SRC collected information, analysed and wrote the manuscript, DG, SRC, HK and SC edited the manuscript; SC arranged the funding.

## Funding

This work was supported by a grant from the Department of Biotechnology, Govt of India vide grant no. BT/PR31648/GET/119/283/2019 to SC. HK acknowledges support from BRNS with 51/14/11/2019-BRNS and SERB India with CRG/2018/001555. DG is supported by the Prime Minister's Research Fellow (PMRF) scholarship.

## Availability of data and materials
Not applicable.

## Ethics approval and consent to participate
Not applicable.

## Consent for publication
All authors have agreed for publication in the journal.
## Competing interests

The authors declare that they have no competing interests.

**Author details**

Molecular Virology Laboratory, School of Life Sciences, Jawaharlal Nehru University, New Delhi-110067, India

Department of Chemical Engineering, Indian Institute of Technology Delhi, New Delhi – 110016, India

School of Artificial Intelligence, Indian Institute of Technology Delhi, New Delhi – 110016, India

**References**

1. Ching T, Himmelstein DS, Beaulieu-Jones BK, Kalinin AA, Do BT, Way GP, Ferrero E, Agapow P-M, Zietz M, Hoffman MM, et al: **Opportunities and obstacles for deep learning in biology and medicine.** 2018, **15:**20170387.
2. Xu C, Jackson SA: **Machine learning and complex biological data.** *Genome Biology* 2019, **20:**76.
3. Zitnik M, Nguyen F, Wang B, Leskovec J, Goldenberg A, Hoffman MM: **Machine learning for integrating data in biology and medicine: Principles, practice, and opportunities.** *Information Fusion* 2019, **50:**71-91.
4. Altman N, Krzywinski M: **The curse(s) of dimensionality.** *Nature Methods* 2018, **15:**399-400.
5. Bzdok D, Altman N, Krzywinski M: **Statistics versus machine learning.** *Nature Methods* 2018, **15:**233-234.
6. Webb SJN: **Deep learning for biology.** 2018, **554:**555-558.
7. LeCun Y, Bengio Y, Hinton G: **Deep learning.** *Nature* 2015, **521:**436-444.
8. Singh A, Ganapathysubramanian B, Singh AK, Sarkar SJTips: **Machine learning for high-throughput stress phenotyping in plants.** 2016, **21:**110-124.
9. Tarca AL, Carey VJ, Chen X-w, Romero R, Drăghici S: **Machine learning and Its applications to biology.** *PLoS Computational Biology* 2007, **3:**e116.
10. Prasad V, Gupta SD: **Applications and potentials of artificial neural networks in plant tissue culture.** In *Plant tissue culture engineering.* Springer; 2008: 47-67
11. Weston J, Leslie C, Ie E, Zhou D, Elisseeff A, Noble WS: **Semi-supervised protein classification using cluster kernels.** *Bioinformatics* 2005, **21:**3241-3247.

12. Tang B, Pan Z, Yin K, Khateeb A: **Recent advances of deep learning in bioinformatics and computational Biology.** 2019, **10**.

13. Mnih V, Kavukcuoglu K, Silver D, Rusu AA, Veness J, Bellemare MG, Graves A, Riedmiller M, Fidjeland AK, Ostrovski G, et al: **Human-level control through deep reinforcement learning.** *Nature* 2015, **518:**529-533.

14. Krenker A, Bešter J, Kos AJANNMA, InTech BA: **Introduction to the artificial neural networks.** 2011**:**1-18.

15. Osama K, Mishra BN, Somvanshi P: **Machine learning techniques in plant biology.** In *PlantOmics: The Omics of plant science.* Springer; 2015: 731-754

16. Yao XJPotI: **Evolving artificial neural networks.** 1999, **87:**1423-1447.

17. Yang ZRJITonn: **A novel radial basis function neural network for discriminant analysis.** 2006, **17:**604-612.

18. Taner A, Öztekin YB, Duran HJS: **Performance analysis of deep learning CNN models for variety classification in hazelnut.** 2021, **13:**6527.

19. Hassan SM, Maji AK, Jasiński M, Leonowicz Z, Jasińska EJE: **Identification of plant-leaf diseases using CNN and transfer-learning approach.** 2021, **10:**1388.

20. Smith LNJapa: **A disciplined approach to neural network hyper-parameters: Part 1--learning rate, batch size, momentum, and weight decay.** 2018.

21. Sarle WSJCs, statistics: **Stopped training and other remedies for overfitting.** 1996**:**352-360.

22. Van der Aalst WM, Rubin V, Verbeek H, van Dongen BF, Kindler E, Günther CWJS, Modeling S: **Process mining: a two-step approach to balance between underfitting and overfitting.** 2010, **9:**87-111.

23. Murakoshi KJB: **Avoiding overfitting in multilayer perceptrons with feeling-of-knowing using self-organizing maps.** 2005, **80:**37-40.

24. Rubio L, Galipienso L, Ferriol I: **Detection of plant viruses and disease management: relevance of genetic diversity and evolution.** 2020, **11**.

25. Varma A, Singh MK: **Chapter 6 - Diagnosis of plant virus diseases.** In *Applied Plant Virology.* Edited by Awasthi LP: Academic Press; 2020: 79-92

26. Bhattacharyya D, Gnanasekaran P, Kumar RK, Kushwaha NK, Sharma VK, Yusuf MA, Chakraborty S: **A geminivirus betasatellite damages the structural and functional integrity of chloroplasts leading to symptom formation and inhibition of photosynthesis.** *Journal of experimental botany* 2015, **66:**5881-5895.

27. Pallas V, García JA: **How do plant viruses induce disease? Interactions and interference with host components.** 2011, **92:**2691-2705.

28. Landgrebe DA: *Signal theory methods in multispectral remote sensing.* John Wiley & Sons; 2003.

29. Lowe A, Harrison N, French AP: **Hyperspectral image analysis techniques for the detection and classification of the early onset of plant disease and stress.** *Plant Methods* 2017, **13:**80.

30. Kawasaki Y, Uga H, Kagiwada S, Iyatomi H: **Basic study of automated diagnosis of viral plant diseases using convolutional neural networks.** In *International symposium on visual computing*. Springer; 2015: 638-645.

31. Ramcharan A, Baranowski K, McCloskey P, Ahmed B, Legg J, Hughes DPJFips: **Deep learning for image-based cassava disease detection.** 2017, **8:**1852.

32. Wang D, Vinson R, Holmes M, Seibel G, Bechar A, Nof S, Tao YJSr: **Early detection of tomato spotted wilt virus by hyperspectral imaging and outlier removal auxiliary classifier generative adversarial nets (OR-AC-GAN).** 2019, **9:**1-14.

33. Kadam KUJIJoFGC, Networking: **Identification of groundnut bud necrosis virus on tomato fruits using machine learning based segmentation algorithm.** 2020, **13:**259-264.

34. Zhu H, Chu B, Zhang C, Liu F, Jiang L, He YJSr: **Hyperspectral imaging for presymptomatic detection of tobacco disease with successive projections algorithm and machine-learning classifiers.** 2017, **7:**1-12.

35. Gu Q, Sheng L, Zhang T, Lu Y, Zhang Z, Zheng K, Hu H, Zhou HJC, Agriculture Ei: **Early detection of tomato spotted wilt virus infection in tobacco using the hyperspectral imaging technique and machine learning algorithms.** 2019, **167:**105066.

36. Griffel L, Delparte D, Edwards JJC, agriculture ei: **Using support vector machines classification to differentiate spectral signatures of potato plants infected with Potato Virus Y.** 2018, **153:**318-324.

37. Joshi RC, Kaushik M, Dutta MK, Srivastava A, Choudhary N: **VirLeafNet: Automatic analysis and viral disease diagnosis using deep-learning in Vigna mungo plant.** *Ecological Informatics* 2021, **61:**101197.

38. Gao Z, Luo Z, Zhang W, Lv Z, Xu YJA: **Deep learning application in plant stress imaging: a review.** 2020, **2:**430-446.

39. Singh AK, Ganapathysubramanian B, Sarkar S, Singh A: **Deep Learning for plant stress phenotyping: trends and future perspectives.** *Trends in Plant Science* 2018, **23:**883-898.

40. Chen Y-M, Zu X-P, Li DJFig: **Identification of proteins of Tobacco mosaic virus by using a method of feature extraction.** 2020, **11:**1186.

41. Mahlein A-K: **Plant disease detection by imaging sensors – parallels and specific demands for precision agriculture and plant phenotyping.** 2016, **100:**241-251.

42. Ferentinos KP: **Deep learning models for plant disease detection and diagnosis.** *Computers and Electronics in Agriculture* 2018, **145:**311-318.

43. Roossinck MJJArog: **Plant virus metagenomics: biodiversity and ecology.** 2012, **46:**359-369.

44. Stobbe AH, Roossinck MJ: **Plant virus metagenomics: what we know and why we need to know more.** 2014, **5**.

45. Li J, Zhang S, Li B, Hu Y, Kang X-P, Wu X-Y, Huang M-T, Li Y-C, Zhao Z-P, Qin C-FJMb, evolution: **Machine learning methods for predicting human-adaptive influenza A viruses based on viral nucleotide compositions.** 2020, **37:**1224-1236.

46. Randhawa GS, Soltysiak MP, El Roz H, de Souza CP, Hill KA, Kari LJPo: **Machine learning using intrinsic genomic signatures for rapid classification of novel pathogens: COVID-19 case study.** 2020, **15:**e0232391.

47. Posada-Céspedes S, Seifert D, Topolsky I, Jablonski KP, Metzner KJ, Beerenwinkel NJB: **V-pipe: a computational pipeline for assessing viral genetic diversity from high-throughput data.** 2021.

48. Elena SF, Agudelo-Romero P, Carrasco P, Codoñer FM, Martín S, Torres-Barceló C, Sanjuán R: **Experimental evolution of plant RNA viruses.** *Heredity* 2008, **100:**478-483.

49. Salama MA, Hassanien AE, Mostafa AJEJoB, Biology S: **The prediction of virus mutation using neural networks and rough set techniques.** 2016, **2016:**1-11.

50. Kumar RV, Singh AK, Singh AK, Yadav T, Basu S, Kushwaha N, Chattopadhyay B, Chakraborty S: **Complexity of begomovirus and betasatellite populations associated with chilli leaf curl disease in India.** 2015, **96:**3143-3158.

51. Devendran R, Kumar M, Ghosh D, Yogindran S, Karim MJ, Chakraborty S: **Capsicum-infecting begomoviruses as global pathogens: host-virus interplay, pathogenesis, and management.** *Trends Microbiol* 2021.

52. Silva JCF, Carvalho TFM, Fontes EPB, Cerqueira FR: **Fangorn Forest (F2): a machine learning approach to classify genes and genera in the family Geminiviridae.** *BMC Bioinformatics* 2017, **18:**431.

53. Mandadi KK, Scholthof K-BG: **Plant immune responses against viruses: how does a virus cause disease?** *The Plant cell* 2013, **25:**1489-1505.

54. Calil IP, Fontes EPB: **Plant immunity against viruses: antiviral immune receptors in focus.** *Ann Bot* 2017, **119:**711-723.

55. Wu X, Valli A, García JA, Zhou X, Cheng X: **The tug-of-war between plants and viruses: great progress and many remaining questions.** *Viruses* 2019, **11**.

56. Kourelis J, van der Hoorn RAL: **Defended to the Nines: 25 years of resistance gene cloning identifies nine mechanisms for R protein function.** *The Plant Cell* 2018, **30:**285-299.

57. Kushwaha SK, Chauhan P, Hedlund K, Ahrén D: **NBSPred: a support vector machine-based high-throughput pipeline for plant resistance protein NBSLRR prediction.** *Bioinformatics* 2015, **32:**1223-1225.

58. Tang D, Wang G, Zhou JM: **Receptor kinases in plant-pathogen interactions: more than pattern recognition.** *Plant Cell* 2017, **29:**618-637.

59. Macho AP, Lozano-Duran R: **Molecular dialogues between viruses and receptor-like kinases in plants.** *Molecular plant pathology* 2019, **20:**1191-1195.

60. Brustolini OJ, Silva JC, Sakamoto T, Fontes EP: **Bioinformatics analysis of the receptor-like kinase (RLK) superfamily.** *Methods Mol Biol* 2017, **1578:**123-132.

61. Liu D, Zhao Q, Cheng Y, Li D, Jiang C, Cheng L, Wang Y, Yang A: **Transcriptome analysis of two cultivars of tobacco in response to Cucumber mosaic virus infection.** *Scientific Reports* 2019, **9:**3124.

62. Liu Y, Liu Y, Spetz C, Li L, Wang X: **Comparative transcriptome analysis in *Triticum aestivum* infecting wheat dwarf virus reveals the effects of viral infection on phytohormone and photosynthesis metabolism pathways.** *Phytopathology Research* 2020, **2:**3.

63. Rajamäki M-L, Sikorskaite-Gudziuniene S, Sarmah N, Varjosalo M, Valkonen JPT: **Nuclear proteome of virus-infected and healthy potato leaves.** *BMC Plant Biology* 2020, **20:**355.

64. Sade D, Shriki O, Cuadros-Inostroza A, Tohge T, Semel Y, Haviv Y, Willmitzer L, Fernie AR, Czosnek H, Brotman Y: **Comparative metabolomics and transcriptomics**

65.   of plant response to Tomato yellow leaf curl virus infection in resistant and susceptible tomato cultivars. *Metabolomics* 2015, **11:**81-97.

65. Mochida K, Koda S, Inoue K, Nishii R: **Statistical and machine learning approaches to predict gene regulatory networks from transcriptome datasets.** 2018, **9**.

66. Rodriguez-Peña R, Mounadi KE, Garcia-Ruiz H: **Changes in subcellular localization of host proteins induced by plant viruses.** *Viruses* 2021, **13:**677.

67. Zhang N, Rao RSP, Salvato F, Havelund JF, Møller IM, Thelen JJ, Xu D: **MU-LOC: a machine-learning method for predicting mitochondrially localized proteins in plants.** *Frontiers in Plant Sciences* 2018, **9:634**

68. Sperschneider J, Catanzariti A-M, DeBoer K, Petre B, Gardiner DM, Singh KB, Dodds PN, Taylor JM: **LOCALIZER: subcellular localization prediction of both plant and effector proteins in the plant cell.** *Scientific Reports* 2017, **7:**44598.

69. Sperschneider J, Gardiner DM, Dodds PN, Tini F, Covarelli L, Singh KB, Manners JM, Taylor JM: **EffectorP: predicting fungal effector proteins from secretomes using machine learning.** *New Phytologist* 2016, **210:**743-761.

70. Ghosh D, M M, Chakraborty S: **Impact of viral silencing suppressors on plant viral synergism: a global agro-economic concern.** *Applied Microbiology and Biotechnology* 2021, **105:**6301-6313.

71. Jagga Z, Gupta D: **Supervised learning classification models for prediction of plant virus encoded RNA silencing suppressors.** *PLoS ONE* 2014, **9:**e97446.

72. Nath A, Subbiah K: **Probing an optimal class distribution for enhancing prediction and feature characterization of plant virus-encoded RNA-silencing suppressors.** *3 Biotech* 2016, **6:**93.

73. Zhang B, Li W, Zhang J, Wang L, Wu J: **Roles of small RNAs in virus-plant interactions.** *Viruses* 2019, **11:**827.

74. Zhang B-T, Nam J-W: **Supervised learning methods for microRNA studies.** In: *Machine learning in bioinformatics* 2008, 339-365

75. Zhang X-M, Liang L, Liu L, Tang M-J: **Graph neural networks and their current applications in bioinformatics.** 2021, **12**.

76. Jumper J, Evans R, Pritzel A, Green T, Figurnov M, Ronneberger O, Tunyasuvunakool K, Bates R, Žídek A, Potapenko A, et al: **Highly accurate protein structure prediction with AlphaFold.** *Nature* 2021, **596:**583-589.

77. Batra R, Chan H, Kamath G, Ramprasad R, Cherukara MJ, Sankaranarayanan SKJTjopcl: **Screening of therapeutic agents for COVID-19 using machine learning and ensemble docking studies.** 2020, **11:**7058-7065.

78. Ma C, Zhang HH, Wang XJTips: **Machine learning for big data analytics in plants.** 2014, **19:**798-808.

79. Ren J, Ahlgren NA, Lu YY, Fuhrman JA, Sun F: **VirFinder: a novel k-mer based tool for identifying viral sequences from assembled metagenomic data.** *Microbiome* 2017, **5:**69.

80. Poland J, Rutkoski J: **Advances and Challenges in Genomic Selection for Disease Resistance.** 2016, **54:**79-98.

81. Crossa J, Pérez-Rodríguez P, Cuevas J, Montesinos-López O, Jarquín D, de los Campos G, Burgueño J, González-Camacho JM, Pérez-Elizalde S, Beyene Y, et al: **Genomic selection in plant breeding: methods, models, and perspectives.** *Trends in Plant Science* 2017, **22:**961-975.

**Figure legends**

**Figure 1. Standard flowchart for creation of a machine learning model to study biological data**. The figure here shows the steps followed in order to create a machine learning model that can successfully study different types of biological data. The data is initially split up into training and testing sets. Each object of the training set is associated with a feature vector, which are passed into the required machine learning algorithm. After manipulating the various parameters of the model, a resultant machine learning model for prediction is developed. This model is then checked by passing the objects of the testing set through it. The resultant output accuracy determines the usefulness of the created model.

**Figure 2. A schematic representation of a standard artificial neural network.** The figure shown here is a standard representation of an artificial neural network. This network is comprised of three basic layers: the input layer, the single hidden layer and the output layer. It is assumed that the input layer has $n$ independent variables, each of which when activated gives a certain output. Depending on this output, the subsequent neurons of the hidden layer. After incorporating the correct bias function, the hidden layer function goes on to activate the output layer, which goes on to produce the final result after considering the bias.

**Figure 3. Application of ML in understanding plant virus pathogenesis.** ML enables early diagnosis of plant viral diseases at field level through analyzing hyperspectral images. Metagenomics study of diseased plant samples helps identification of related and unrelated viral genomes. ML can assist classification of these viral sequences which primes our understanding of virus evolution. Furthermore, ML assisted bioinformatics tools have been developed to identify viral suppressors of RNA silencing (VSRs). ML can also guide us to predict the sub-cellular localization and even the structure of the viral proteins. Prediction of accurate structure of virus encoded proteins may help to identify inhibitors of these effector proteins. To understand the host response, several groups have performed transcriptome, proteome and metabolome of virus infected plants. ML can prime the accurate and fast analysis of these high throughput data to identify gene regulatory networks (GRN) and novel host factors involved in host-virus interplay. Characterization of these host factors in terms of sub-cellular localization and structure prediction will boost understanding of plant virus pathogenesis. ML may also assist plant virologists in genomic selection to identify elite virus resistant cultivars. This figure was created using BioRender (https://biorender.com/).

**Table 1. The application of ML assisted diagnosis of plant viral diseases.**

**Figure 1**

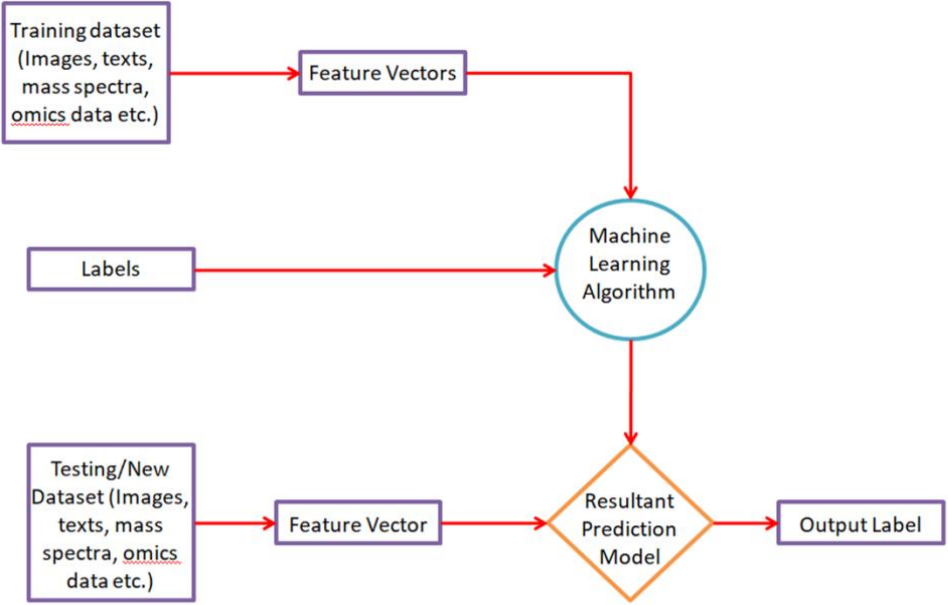

**Figure 2**

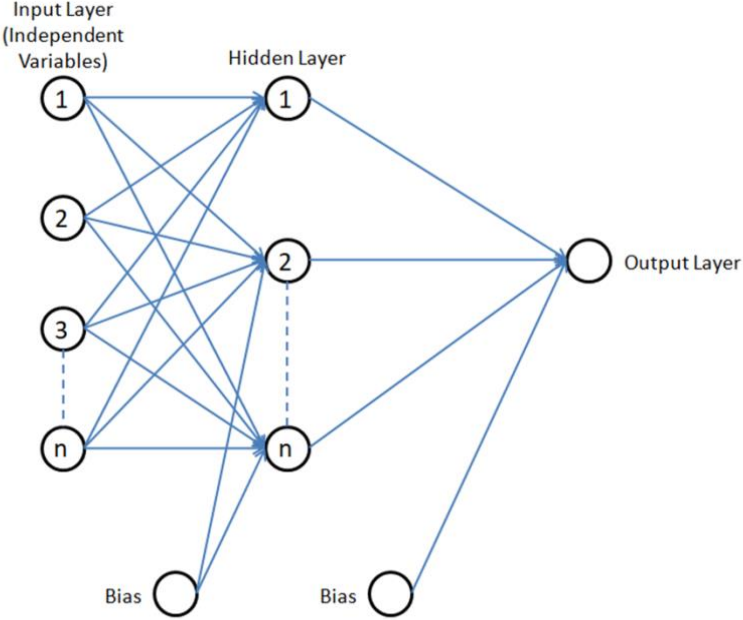

**Figure 3**

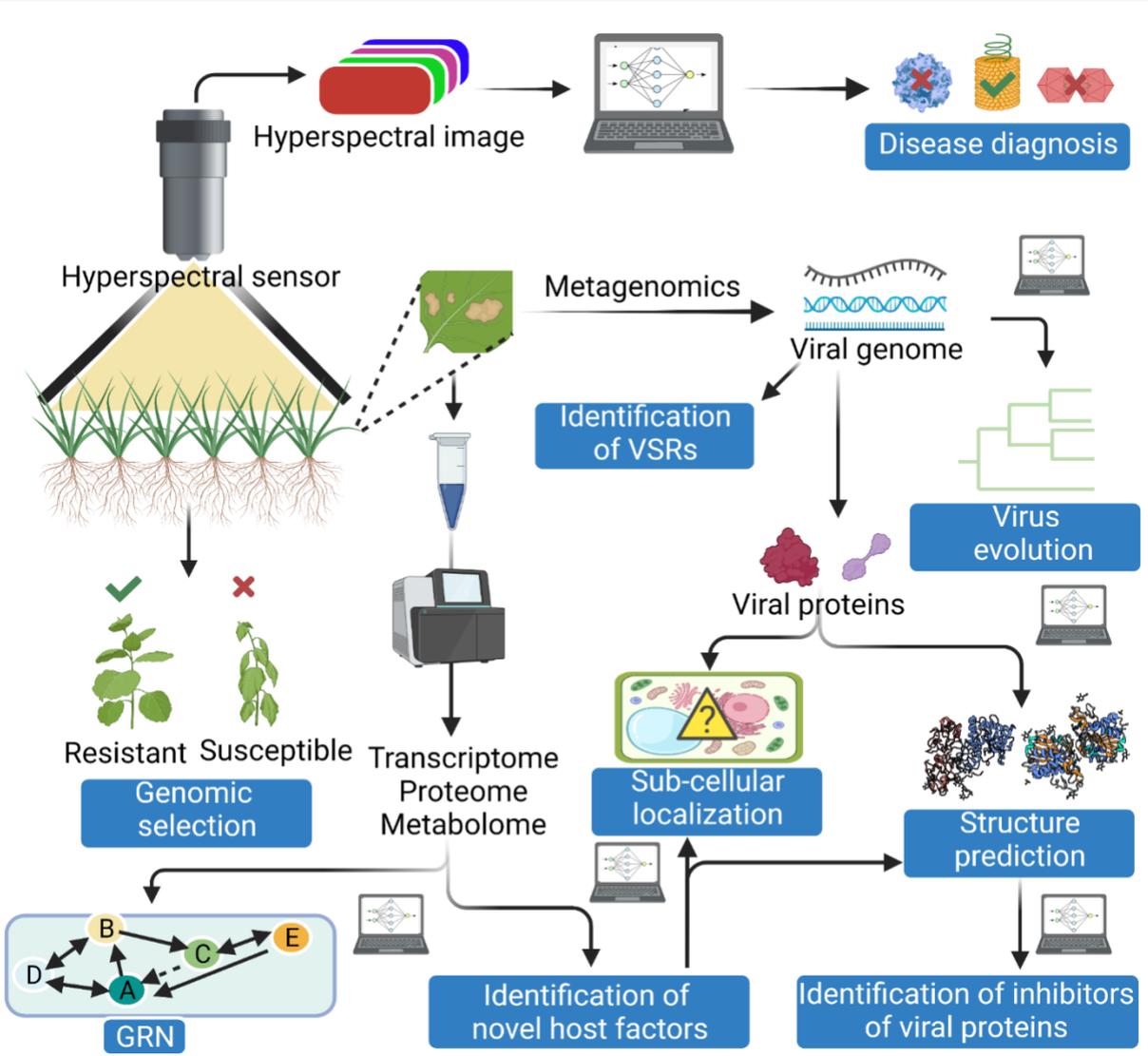
23